\documentclass[runningheads]{llncs}
\usepackage{graphicx}
\usepackage{amsmath,amssymb,amsfonts}
\usepackage{algorithmic}
\usepackage{graphicx}
\usepackage{textcomp}
\usepackage{xcolor}
\usepackage{subfigure}
\usepackage{color,xcolor,soul,colortbl}
\usepackage{caption}
\usepackage{wrapfig}
\begin{document}
\title{``iCub! Do you recognize what I am doing?": multimodal human action recognition on multisensory-enabled iCub robot}
%
\author{Kas Kniesmeijer\inst{1} \and
Murat Kirtay\inst{2}}

%
\authorrunning{Kniesmeijer et al.}
%
\institute{Data Science and Society, Tilburg University, Tilburg, The Netherlands \\ 
\email{\{k.l.kniesmeijer\}@tilburguniversity.edu}\and
Department of Cognitive Science and Artificial Intelligence, Tilburg University, Tilburg, The Netherlands\\
\email{\{m.kirtay\}@tilburguniversity.edu}}
\maketitle              
\begin{abstract}

This study uses multisensory data (i.e., color and depth) to recognize human actions in the context of multimodal human-robot interaction. Here we employed the iCub robot to observe the predefined actions of the human partners by using four different tools on 20 objects. We show that the proposed multimodal ensemble learning leverages complementary characteristics of three color cameras and one depth sensor that improves, in most cases, recognition accuracy compared to the models trained with a single modality. The results indicate that the proposed models can be deployed on the iCub robot that requires multimodal action recognition, including social tasks such as partner-specific adaptation, and contextual behavior understanding, to mention a few.

\keywords{multimodal learning  \and action recognition \and multimodal HRI.}
\end{abstract}
\section{Introduction}

Action recognition is a critical skill for achieving collaborative tasks (such as working with a co-bot to assemble furniture) in human-robot interaction \cite{ji2019survey}. To recognize human partners' actions, the robot should learn the consequence of the action performed on an object by employing its sensors. The computational models of recognizing human actions can be framed in the context of internal model learning, affordance learning, and multimodal machine learning \cite{escobar2016self}, \cite{jamone2016affordances},  \cite{Ramachandram2017survey}. Here we follow a multimodal machine learning approach to recognize human partners’ actions. In this study, we create a human-robot interaction scenario in which human operators are action performers who use a set of tools on the objects. To this end, we employed the iCub robot as an action observer to record objects' initial and final positions using three color cameras and a depth sensor. For the purpose of this study, we adopt the definition of modalities as described by Latah et al, where phenomena are observed by using multiple sensors and each sensor can be termed a modality associated with a single data set \cite{lahat2015multimodal}. After acquiring data, we develop multiple machine learning models to recognize human actions. To assess the performance of the action recognition model, we designed two different settings: unimodal and multimodal. In the unimodal setting, we formed convolutional neural network-based architectures for each sensor. In the multimodal setting, we combined the prediction of each sensor to build an ensemble learning for the same problem. The preliminary results show that, in most cases, employing multimodal data improves recognition accuracy compared to unimodal one.

This study offers the following contributions. On the one hand, we provide multisensory learning results of human action recognition using a novel dataset constructed using the iCub robot. On the other hand, we provide the first results of the iCub robot dataset that other researchers could leverage in this study to benchmark their models. 


\section{Experimental setup and dataset specifications} \label{dataset}

This section introduces the experimental setup and the dataset acquisition procedures~\cite{kirtay2020icub2}. The experimental setup consists of the iCub robot with two color cameras (Dragonfly) mounted in the robot’s eyes and an additional Intel Realsense d435i camera located above the eyes. 
\begin{figure}[ht!]
	\centering
	\begin{center}
 		\subfigure[Objects ]{\label{toolsAndObjs:a}\includegraphics[width=0.28\textwidth, height=0.15\textheight]{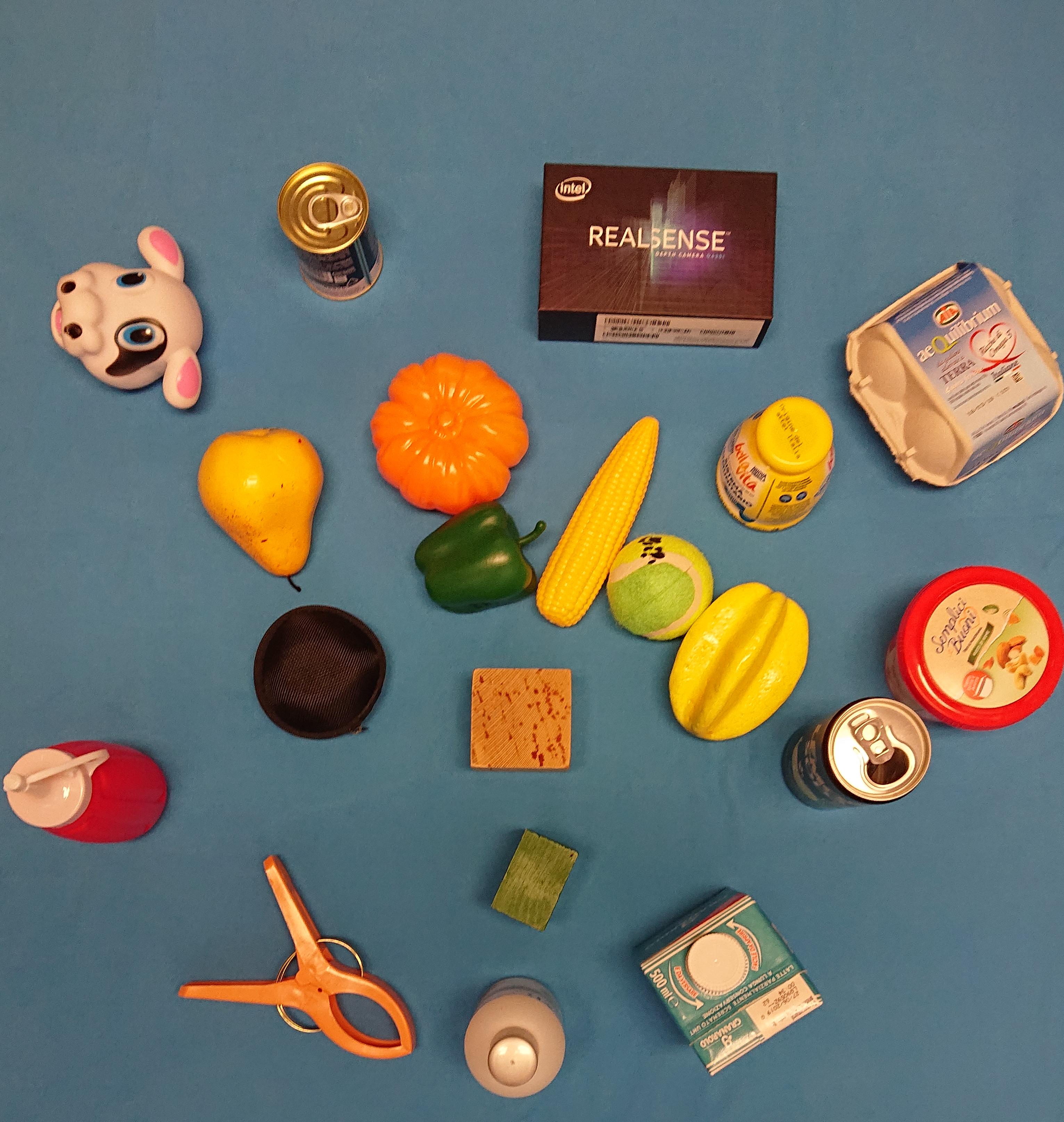}} 
		\subfigure[Tools ]{\label{toolsAndObjs:b}\includegraphics[width=0.28\textwidth, height=0.15\textheight]{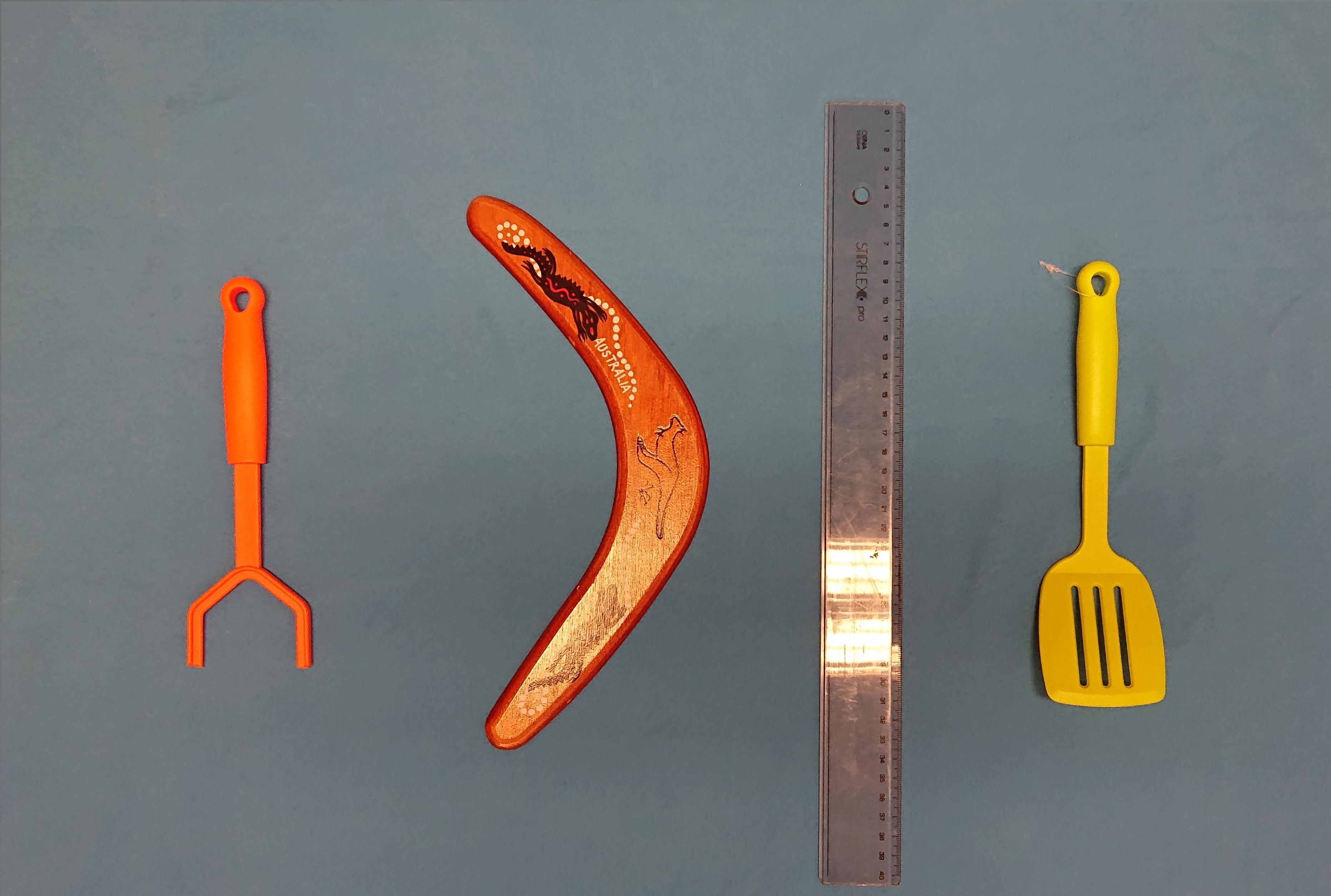}}
		\subfigure[Experiment setup]{\label{toolsAndObjs:c}\includegraphics[width=0.42\textwidth]{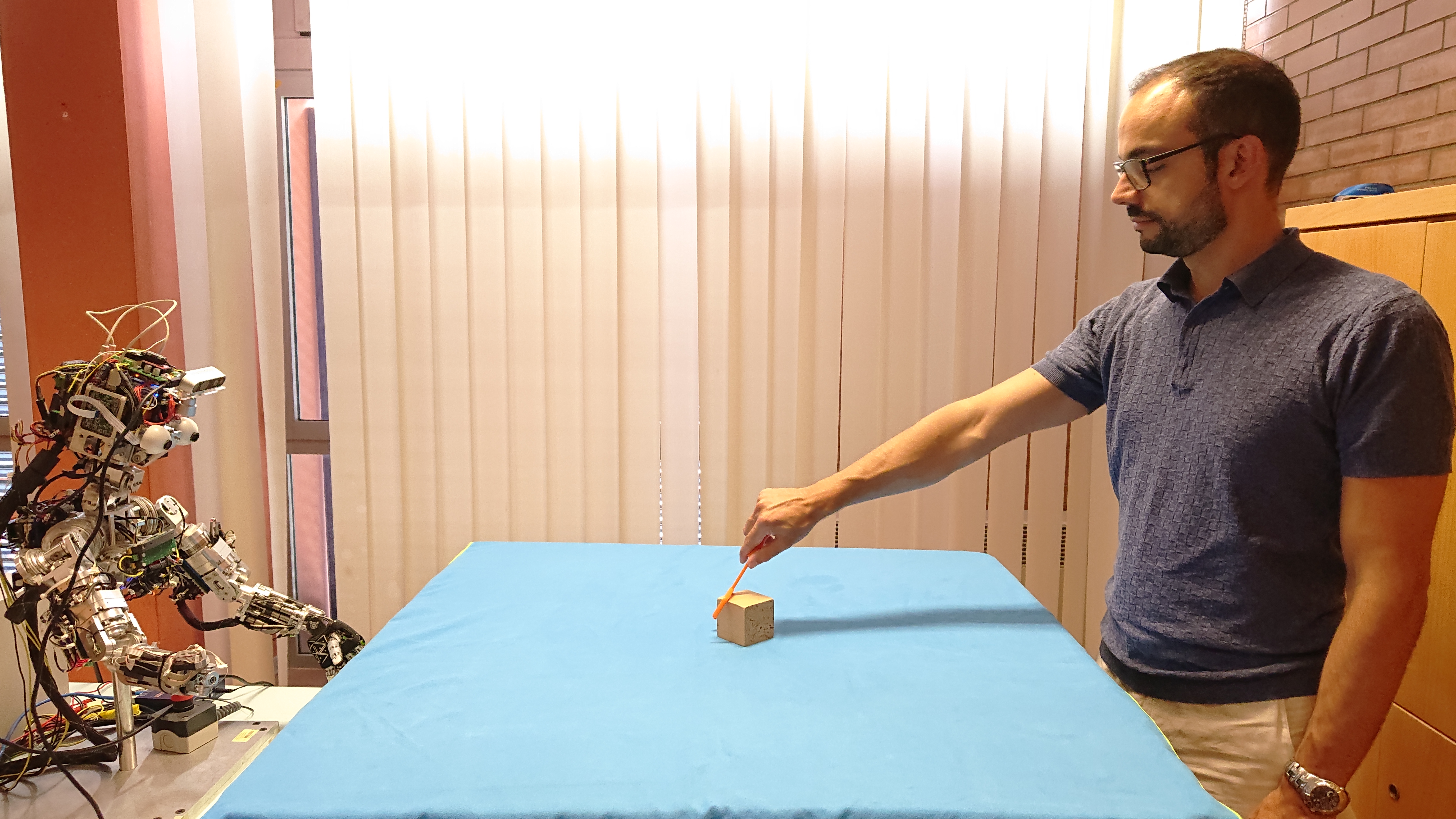}}\\

  		\subfigure[Push ]{\label{actions:a}\includegraphics[width=0.23\textwidth, height=0.15\textheight]{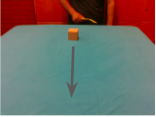}} 
		\subfigure[Pull ]{\label{actions:b}\includegraphics[width=0.23\textwidth, height=0.15\textheight]{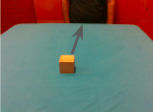}}
		\subfigure[Left to right]{\label{actions:c}\includegraphics[width=0.23\textwidth, height=0.15\textheight]{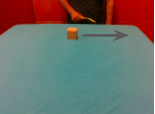}}
         \subfigure[Right to left]{\label{actions:d}\includegraphics[width=0.23\textwidth, height=0.15\textheight]{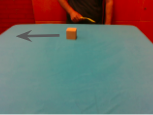}}

	\end{center}
	\caption{ The first row: The objects (a), the tools (b), and the experimental setup (c) that were employed to construct the dataset. The second row shows the operator actions.}
	\label{fig:toolsAndObjs}
\end{figure}
By capturing data in a multisensory setting, we aimed to test the complementary characteristics of different sensors in performing human action recognition. 

In the interactive experiments, we consider the iCub robot as an action observer and four human operators as action performers. In this setting, each operator performs four predefined actions on the objects by using four different tools. The dataset was constructed with 20 objects with specific affordances, shown in Figure~\ref{toolsAndObjs:a}. For instance, a can of liquid drink offers rollability as affordance, whereas a box can not provide the same affordance \cite{jamone2016affordances}. The set of tools used in the experiments is presented in Figure~\ref{toolsAndObjs:b}. 

As seen in Figure~\ref{toolsAndObjs:c}, the iCub robot is situated in front of the operator to capture the images with three color cameras and a depth sensor. We note that the images were recorded before and after performing actions. In doing so, we obtained the pose of the object in the scene and the effect of action (i.e., final pose) that occurred by using a specific tool to perform a particular action. We allow the operators to perform four different actions: pulling, pushing, moving an object from left to right, and moving an object from right to left.  The subfigures in the second row of Figure~\ref{fig:toolsAndObjs} were captured from the robot's camera, and the arrows illustrate the possible trajectory of an object with these actions. We repeat the same action 10 times for each object-tool-action triplet and collected 9600 color and 3200 depth images (with the size of   $640\times480$) of the initial and final poses of the object.


\section{Action recognition pipeline} \label{pipeline}

This section presents the action recognition pipeline on the iCub robot dataset. To process image data, we first downsized the images to $64\times64$.  Then, the pixel values were normalized, and the matrices flattened into vectors. Next, the initial and final pose vectors were concatenated to form the inputs. Here, the corresponding action labels are outputs of the models. This process was repeated for all sensory data. Finally, the input vectors were randomly grouped into train, validation, and test datasets using 60\%, 20\%, and 20\% of the dataset.

To recognize the actions, we employed three models. These models are a Multilayer Perceptron (MLP) and two types of Convolutional Neural Networks (CNN-1 and CNN-2). All models were built and tuned on IntelRealsense’s color images (hereafter, Color dataset), and the CNN-1 model was used as a benchmark model to measure performance improvement of the CNN-2 model. To tune hyperparameters, we used a grid search for MLP models and a random search for CNN-2. The MLP model has three hidden layers with 120, 80, and 40 units, respectively, and \emph{tanh} as an activation function. The models were trained to minimize categorical cross-entropy loss, and weights were updated using stochastic gradient descent as a solver. These models were trained for 500 epochs.

The CNN-1 model consists of three convolutional layers with  \emph{ReLU} as an activation function. These convolutional layers have 32, 64 and 128 output filters in the convolution, respectively. After a convolutional layer, a max pooling layer was added to downsample the input. After standard CNN operations, we flattened the filters to be an input of a fully-connected network with 128 nodes. Finally, the output layer hosts 4 nodes --which corresponds with the four actions performed by the operator-- with the Softmax function. Here the Adam optimizer was used, and categorical cross-entropy was the loss function. The model was then trained on the training set using 20 epochs. The CNN-2 model has the same convolutional and pooling layers as CNN-1. In addition, it has dropout layers with a rate of 0.2 between the convolutional and hidden layers. The CNN-2 model has three hidden layers with 480, 224, and 32 nodes, respectively. The first hidden layer uses the Sigmoid activation function, and the other uses the ReLU. The output layers, number of the epoch, and training procedures have the same characteristics as the CNN-1.
\begin{wrapfigure}{r}{0.45\textwidth} 
  \centering
  \begin{center}
  \includegraphics[width=0.45\textwidth, height=0.2\textheight]{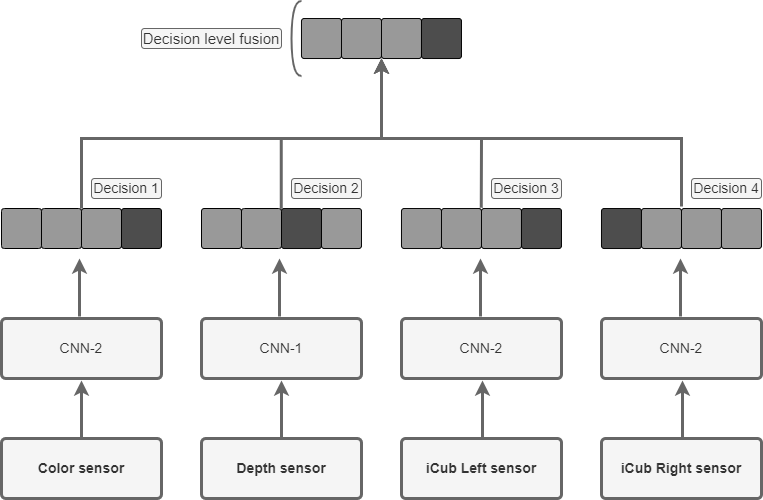}
  \end{center}
  \caption{Decision level fusion architecture for action recognition}
  \label{fig:dlf}
\end{wrapfigure}
After training the models, we selected the best-performing models on each sensor for decision-level fusion, shown in Figure \ref{fig:dlf}. Three types of decision-level fusion are described. At first, we took the average four models' predictions. Secondly, we added coefficients to each model to improve the performance. Lastly, we demonstrate how weights found by post-processing data allow for gaining insight into the full potential of decision-level fusion for the dataset.  For average decision-level fusion, the method is applied in the following way: first, the best model per sensor is selected. Then, each model predicts the class for each instance in the test set. After that, the probabilities of a class are added up and normalized.  Then, the class with the highest probability is chosen by the decision-level fusion model. 
The weighted average decision level fusion method was applied in the following way. At first, the best model per sensor is selected. Then, coefficients (weights) are added to each model. By adding coefficients to the models, we aimed to influence each model's contribution to action recognition. For this method, a function was employed, randomly assigning coefficients between 0-1. Then, the combination of coefficients that yielded the highest accuracy on the validation set was selected. This process was repeated 10 times, and the results were averaged. The averaged results then became the coefficients for the models. Thereafter, the weighted probabilities of an instance belonging to a class are added to obtain the weighted combined probability of an instance belonging to a class. Then, the class with the highest probability is chosen by the decision-level fusion. 

\section{Results} \label{results}

\begin{figure}[ht!]
	\centering
	\begin{center}
 		\subfigure[Average]{\label{decision level fusion:a}\includegraphics[width=0.32\textwidth, height=0.16\textheight]{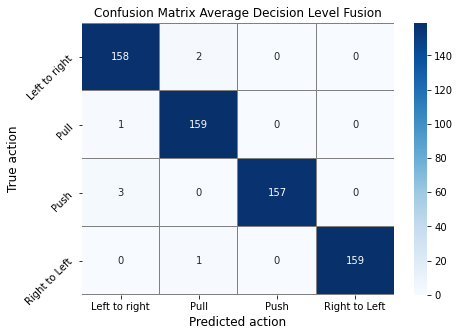}} 
		\subfigure[Weighted average]{\label{decision level fusion:b}\includegraphics[width=0.32\textwidth, height=0.16\textheight]{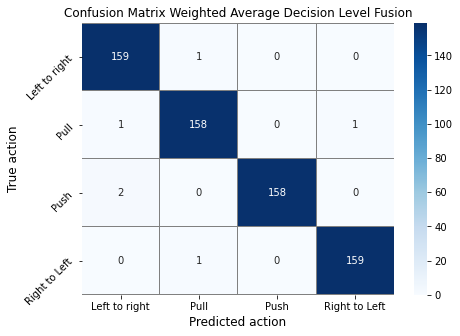}}
		\subfigure[Weighted average PP]{\label{decision level fusion:c}\includegraphics[width=0.32\textwidth, height=0.16\textheight]{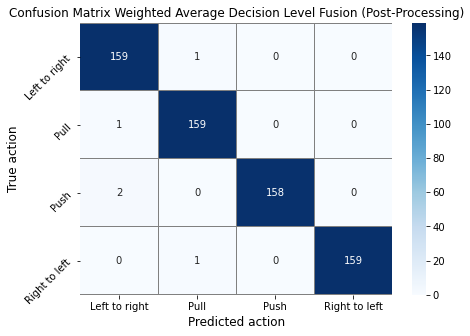}}
	\end{center}
	\caption{ The confusion matrices of the decision level fusion methods (a), Average (b) Weighted average (c) Weighted average PP (PP indicates post-processing).}
	\label{fig:decision level fusion}
\end{figure}

This section introduces the results obtained from the test set. Based on the entries in Table \ref{t:performance}, for the Color, iCub Left, and iCub Right sensors, the best-performing model is the CNN-2 model. For the Depth sensor, the best-performing model is the CNN-1 model. For decision level fusion, the resulting coefficients for weighted average from the method described in Section \ref{pipeline} are: 0.2628, 0.1057, 0.2888, 0.3427 for the Color, Depth, iCub Left and iCub Right models, respectively. Table \ref{t:performance} also shows the evaluation metrics of the different decision-level fusion types compared to the model's single modality versions. The coefficients applied to the result from the method on post-processing data are: 0.40, 0.05, 0.15, 0.40 for the Color, Depth, iCub Left, and iCub Right models. For the simple average method, no coefficients were applied. Figure \ref{fig:decision level fusion} shows the confusion matrices of the three decision-level fusion methods.

\begin{table}[ht!]
    \begin{center}
        \begin{tabular}{ccccc}& \multicolumn{4}{c}{\textbf{Performance Metrics}} \\ \cline{2-5} 
            \textbf{Sensors}                        & \textit{Accuracy}           & \textit{Precision}          & \textit{ Recall }             & \textit{$F_1$ score}        \\ \hline
            \multicolumn{1}{|c|}{Color sensor}      & \multicolumn{1}{c|}{0.9906} & \multicolumn{1}{c|}{0.9907} & \multicolumn{1}{c|}{0.9906} & \multicolumn{1}{c|}{0.9906} \\ \hline
            \multicolumn{1}{|c|}{Depth sensor}      & \multicolumn{1}{c|}{0.9531} & \multicolumn{1}{c|}{0.9548} & \multicolumn{1}{c|}{0.9531} & \multicolumn{1}{c|}{0.9533} \\ \hline
            \multicolumn{1}{|c|}{iCub Left sensor}  & \multicolumn{1}{c|}{0.9859} & \multicolumn{1}{c|}{0.9861} & \multicolumn{1}{c|}{0.9859} & \multicolumn{1}{c|}{0.9860} \\ \hline
            \multicolumn{1}{|c|}{iCub Right sensor} & \multicolumn{1}{c|}{0.9875} & \multicolumn{1}{c|}{0.9876} & \multicolumn{1}{c|}{0.9875} & \multicolumn{1}{c|}{0.9875} \\ \hline
            \multicolumn{1}{|c|}{Average DLF}       & \multicolumn{1}{c|}{0.9891} & \multicolumn{1}{c|}{0.9892} & \multicolumn{1}{c|}{0.9891} & \multicolumn{1}{c|}{0.9891} \\ \hline
            \multicolumn{1}{|c|}{\textbf{Weighted average DLF}} & \multicolumn{1}{c|}{\textbf{0.9906}} & \multicolumn{1}{c|}{\textbf{0.9907}} & \multicolumn{1}{c|}{\textbf{0.9906}} & \multicolumn{1}{c|}{\textbf{0.9906}} \\ \hline
            \multicolumn{1}{|c|}{Weighted average DLF (post-processing)} & \multicolumn{1}{c|}{0.9922} & \multicolumn{1}{c|}{0.9923} & \multicolumn{1}{c|}{0.9922} & \multicolumn{1}{c|}{0.9922} \\ \hline 
        \end{tabular}
    \end{center}
    \caption{ Performances decision level fusion methods compared to single modality versions of the models. In this table ‘decision level fusion’ is abbreviated to ‘DLF’.}
    \label{t:performance}
\end{table}

As seen from Table \ref{t:performance}, the average decision level fusion has an accuracy of 0.9891, a precision of 0.9892, a recall of 0.9891, and an F1-score of 0.9891. This is slightly less than that of the Color sensor model (a decrease of 0.0015 across all metrics). However, it performs better than three out of four single-modality models. Namely, the Depth sensor model (0.9531 accuracy), the iCub Left sensor model (0.9859 accuracy), and the iCub Right sensor model (0.9875 accuracy). This means an improvement of 0.0360, 0.0032, and 0.0016, respectively. 
The weighted average decision level fusion from Table \ref{t:performance} (highlighted in Bold) has an accuracy of 0.9906, a precision of 0.9907, a recall of 0.9906, and an F1-score of 0.9906. These are the same performances as the Color sensor model, but also an increase in performance from the average decision level fusion. 

Therefore, decision-making was influenced by adding weights to the models; in turn, it improved the accuracy by 0.0015. That translates to an increase of 1 true positive. This can also be seen in the corresponding confusion matrices from Figure \ref{fig:decision level fusion}. Although the results given by the post-processing weighted average decision level fusion might not be achieved in an online action recognition case, this approach gives us insight into the full potential of decision level fusion for this particular dataset. Table \ref{t:performance} shows an accuracy of 0.9922, a precision of 0.9923, a recall of 0.9922, and an F1-score of 0.9922 for this ensemble learning type. This indicates that the selected model weights did not achieve the full learning potential on this dataset. To reproduce the presented results and provide all the related data to the interested researchers, we used a public repository~\footnote{https://github.com/KasKniesmeijer/iCub-Multimodal-HRI}. We note that the same repository also hosts the results with MLP models that we could not provide here due to the page limit.

\section{Conclusions}\label{conc}
In this study, we have demonstrated multimodal human action recognition using the iCub robot as an action observer. The results indicate that leveraging multisensory data with ensemble learning often improves action recognition performance compared to the models trained with a single modality (color or depth) as inputs. In addition, the presented results can be seen as preliminary results toward learning human affordances (i.e., action-tool-effect triplets, we provide only action recognition results in this paper, though) by the iCub robot.

%

%
%
%
\bibliographystyle{splncs04}
\bibliography{muratCSAI}

\begin{thebibliography}{1}
\providecommand{\url}[1]{\texttt{#1}}
\providecommand{\urlprefix}{URL }
\providecommand{\doi}[1]{https://doi.org/#1}

\bibitem{escobar2016self}
Escobar-Ju{\'a}rez, E., Schillaci, G., Hermosillo-Valadez, J., Lara-Guzm{\'a}n,
  B.: A self-organized internal models architecture for coding sensory--motor
  schemes. Frontiers in Robotics and AI  \textbf{3}, ~22 (2016)

\bibitem{jamone2016affordances}
Jamone, L., Ugur, E., Cangelosi, A., Fadiga, L., Bernardino, A., Piater, J.,
  Santos-Victor, J.: Affordances in psychology, neuroscience, and robotics: A
  survey. IEEE Transactions on Cognitive and Developmental Systems
  \textbf{10}(1),  4--25 (2016)

\bibitem{ji2019survey}
Ji, Y., Yang, Y., Shen, F., Shen, H.T., Li, X.: A survey of human action
  analysis in hri applications. IEEE Transactions on Circuits and Systems for
  Video Technology  \textbf{30}(7),  2114--2128 (2019)

\bibitem{kirtay2020icub2}
Kirtay, M., Albanese, U., Vannucci, L., Schillaci, G., Laschi, C., Falotico,
  E.: The icub multisensor datasets for robot and computer vision applications.
  In: Proceedings of the 2020 International Conference on Multimodal
  Interaction. pp. 685--688 (2020)

\bibitem{lahat2015multimodal}
Lahat, D., Adali, T., Jutten, C.: Multimodal data fusion: an overview of
  methods, challenges, and prospects. Proceedings of the IEEE  \textbf{103}(9),
   1449--1477 (2015)

\bibitem{Ramachandram2017survey}
{Ramachandram}, D., {Taylor}, G.W.: Deep multimodal learning: A survey on
  recent advances and trends. IEEE Signal Processing Magazine  \textbf{34}(6),
  96--108 (Nov 2017). \doi{10.1109/MSP.2017.2738401}

\end{thebibliography}

\end{document}